\documentclass{article}

\usepackage[preprint]{neurips_2026}

\usepackage[utf8]{inputenc}
\usepackage[T1]{fontenc}
\usepackage{hyperref}
\usepackage{url}
\usepackage{booktabs}
\usepackage{multirow}
\usepackage{amsmath}
\usepackage{amsfonts}
\usepackage{nicefrac}
\usepackage{microtype}
\usepackage[table]{xcolor}
\usepackage{graphicx}
\usepackage{makecell}
\usepackage{wrapfig}
\usepackage[most]{tcolorbox}
\usepackage{fvextra}
\usepackage{enumitem}
\usepackage{underscore}
\fvset{fontsize=\footnotesize, baselinestretch=1.05, breaklines=true, breakanywhere=true, breaksymbolleft={}, breaksymbolright={}}

\newtcolorbox{promptbox}[1][]{%
  enhanced,
  breakable,
  colback=gray!3,
  colframe=black!55,
  boxrule=0.4pt,
  arc=2pt,
  left=10pt, right=10pt, top=6pt, bottom=6pt,
  before skip=6pt, after skip=10pt,
  #1
}

\title{Faithful-MR1: Faithful Multimodal Reasoning via Anchoring and Reinforcing Visual Attention}

\author{%
  \textbf{Changyuan Tian}\textsuperscript{1,2,*} \quad
  \textbf{Zhicong Lu}\textsuperscript{2,*,\textdagger} \quad
  \textbf{Huaxing Liu}\textsuperscript{1} \quad
  \textbf{Xiang Wang}\textsuperscript{1,\textdaggerdbl} \\
  \textbf{Shuai Li}\textsuperscript{1} \quad
  \textbf{Yu Chen}\textsuperscript{1} \quad
  \textbf{Wenqian Lv}\textsuperscript{1} \quad
  \textbf{Zichuan Lin}\textsuperscript{3} \quad
  \textbf{Juncheng Diao}\textsuperscript{2} \quad
  \textbf{Deheng Ye}\textsuperscript{4} \\[4pt]
  \textsuperscript{1}AMAP, Alibaba Group \quad
  \textsuperscript{2}University of Chinese Academy of Sciences \\
  \textsuperscript{3}Tsinghua University \quad
  \textsuperscript{4}Nanyang Technological University
}

\begin{document}

\maketitle

\begingroup
\renewcommand\thefootnote{}%
\footnotetext{\textsuperscript{*}Equal contribution.\quad \textsuperscript{\textdagger}Corresponding author.\quad \textsuperscript{\textdaggerdbl}Project leader.\quad Work done during internship at AMAP, Alibaba Group.\quad Contact: \texttt{tianchangyuan21@mails.ucas.edu.cn}, \texttt{nazaritelzc@gmail.com}.}%
\endgroup

\begin{abstract}
Reinforcement learning with verifiable rewards (RLVR) has emerged as a promising paradigm for advancing complex reasoning in large language models, and recent work extends RLVR to multimodal large language models (MLLMs). This transfer, however, surfaces a faithfulness challenge: faithful perception of task-relevant visual evidence and faithful use of that evidence during reasoning, leading to unsatisfactory gains on multimodal benchmarks. Specifically, existing perception supervision often operates on textual descriptions rather than natively on image regions, and faithful use is largely overlooked, exposing the \emph{perception--reasoning disconnect} where correctly perceived evidence is dropped or contradicted during reasoning. To close these gaps, we propose \textbf{Faithful-MR1}, a training framework that anchors and reinforces visual attention to address both halves of faithful multimodal reasoning. The {Anchoring stage} turns perception into an explicit pre-reasoning subtask, supervising a dedicated \texttt{<Focus>} token's attention directly against image regions rather than through textual descriptions. The {Reinforcing stage} exposes faithful use through counterfactual image intervention, rewarding answer-correct trajectories that concentrate visual attention where vision causally matters. Extensive experiments demonstrate that Faithful-MR1 outperforms recent multimodal reasoning baselines on both Qwen2.5-VL-Instruct 3B and 7B backbones while using substantially less training data.
\end{abstract}

\section{Introduction}
\label{sec:intro}

Reinforcement learning with verifiable rewards (RLVR) has emerged as a promising paradigm for advancing complex reasoning in large language models~\citep{DeepSeek-R1,GRPO,jaech2024openai,team2025kimi,lu2026hisr}. Motivated by this success, a growing line of work extends RLVR to multimodal large language models (MLLMs), aiming to bring similar reasoning improvements to vision-language tasks~\citep{VisionR1,MM-Eureka,R1-V,Vlm-r1,Visual-rft,yang2025r1,xu2025llava}.

On multimodal reasoning benchmarks, however, this transferred recipe yields substantially smaller gains than in the text-only setting. Unlike text-only reasoning, a multimodal reasoning problem requires both faithful \emph{perception} of task-relevant visual evidence from the image and faithful \emph{use} of that evidence in the reasoning chain. Recent work addresses parts of these requirements: Perception-R1~\citep{PerceptionR1} and Vision-SR1~\citep{VisionSR1} introduce supervision that rewards faithful visual descriptions, and VPPO~\citep{VPPO} reweights policy updates toward visually pivotal tokens. Despite these advances, two underlying gaps cut across these efforts. First, perception is supervised through generated text, an indirect signal for how the model extracts task-relevant visual evidence native to the image. Second, faithful use is overlooked, breaking the link from perception to reasoning: even when task-relevant visual evidence is correctly perceived, it may still be dropped, distorted, or overridden by language priors in the reasoning chain---a failure mode we term the \emph{perception--reasoning disconnect} (PRD).

\begin{figure}[!t]
  \centering
  \includegraphics[width=\linewidth]{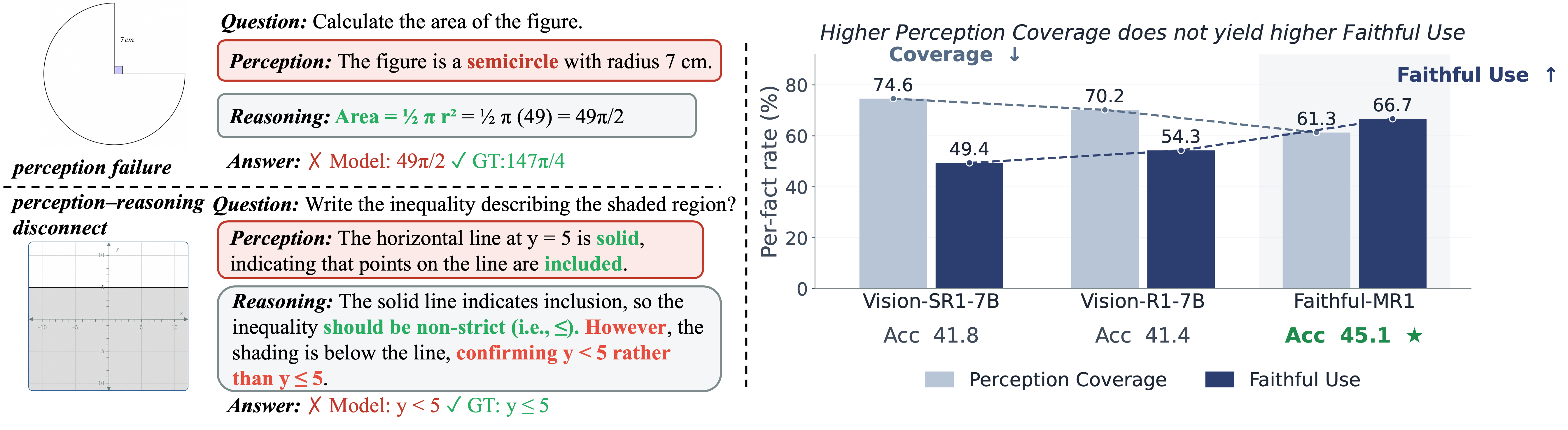}
  \caption{
    Two failure modes in current multimodal RLVR.
    \textbf{Top-left}: \emph{perception failure}---the model misreads
    visual evidence, and the chain compounds the error.
    \textbf{Bottom-left}: \emph{perception--reasoning disconnect}
    (PRD)---visual evidence is correctly perceived but contradicted
    during the derivation.
    \textbf{Right}: as Perception Coverage decreases across methods,
    Faithful Use \emph{rises} and tracks final answer accuracy
    (Section~\ref{sec:prd_analysis}); Faithful-MR1 leads on both.
  }
  \label{fig:failure_modes}
\end{figure}

To address these gaps, we conduct a focused quantitative analysis (Figure~\ref{fig:failure_modes}) of two coexisting failure modes: a \emph{perception failure}, where the chain starts from incorrect visual evidence, and a \emph{PRD failure}, where the evidence is correctly perceived but later dropped or contradicted. Across these methods, faithful perception alone does not necessarily translate to faithful use, and final answer accuracy appears more directly influenced by faithful use than by faithful perception alone (Section~\ref{sec:prd_analysis}). PRD is therefore a distinct and quantitatively significant failure mode, not subsumed by perception accuracy alone.

Building on this analysis, we propose \textbf{Faithful-MR1}, a training framework that anchors and reinforces visual attention to address both halves of faithful multimodal reasoning: faithful \emph{perception} of task-relevant visual evidence from the image, and faithful \emph{use} of that evidence in the reasoning chain. The \textbf{Anchoring stage} turns perception into an explicit one-step pre-reasoning subtask (``focus first, then reason''): before producing the full reasoning trajectory, the model must first localize and gather the relevant visual evidence through a dedicated \texttt{<Focus>} token, whose attention is supervised directly against image regions rather than generated text. The \textbf{Reinforcing stage} faces a subtler challenge: faithful use is not directly observable in the reasoning trace. We expose it through counterfactual image intervention, treating response tokens whose predictions shift under image masking as where vision causally matters, and reward answer-correct trajectories that concentrate visual attention there. Faithful-MR1 realizes faithful multimodal reasoning---perception is grounded in the image, and reasoning is grounded in perception.

Our contributions are:
\begin{itemize}[leftmargin=1.6em, itemsep=2pt, topsep=3pt]
  \item Through a focused quantitative analysis (Section~\ref{sec:prd_analysis}), we reveal that pursuing more accurate perception text does not translate to higher answer accuracy in current multimodal RLVR, exposing the perception--reasoning disconnect as a distinct failure mode beyond the reach of perception supervision alone.
  \item We propose \textbf{Faithful-MR1}, a training framework for faithful multimodal reasoning that couples an Anchoring stage with a Reinforcing stage, jointly closing the perception and use gaps.
  \item We construct an evidence-grounded training corpus of 19.2K examples with bounding-box annotations of question-relevant regions, providing a reusable resource for region-grounded supervision in multimodal RLVR; we will release the corpus, training code, and trained checkpoints to facilitate further research.
  \item With substantially less training data, Faithful-MR1 outperforms current multimodal RLVR baselines across the Qwen2.5-VL-Instruct 3B and 7B backbones, and additional analysis confirms substantive improvements in faithful multimodal reasoning.
\end{itemize}

\section{Related Work}

\subsection{RLVR for Multimodal Reasoning}

Reinforcement Learning with Verifiable Rewards (RLVR) has become a powerful paradigm for improving language-model reasoning in domains where final answers can be checked automatically~\citep{DeepSeek-R1,jaech2024openai,team2025kimi,tian2026rectify}. RLVR is typically instantiated on top of PPO~\citep{schulman2017proximal}; its group-relative variant GRPO~\citep{GRPO} replaces per-sample value estimation with group-based advantage normalization, and DAPO~\citep{yu2025dapo} further refines GRPO with techniques that stabilize training at scale. Recent work adapts this paradigm to multimodal large language models (MLLMs) by defining verifiable rewards over image-text tasks. Vision-R1~\citep{VisionR1} demonstrates the effectiveness of R1-style answer-level RLVR for multimodal reasoning, and a broad line of follow-up work explores rule-based RL across diverse task families, training scales, rollout strategies, and shortcut-mitigation designs~\citep{MM-Eureka,R1-V,Vlm-r1,Visual-rft,yang2025r1,wang2025skywork,liu2025noisyrollout,xia2025visionary}. A parallel direction scales long-chain visual reasoning through structured supervision~\citep{xu2025llava,thawakar2025llamav}. These methods establish outcome-verifiable reinforcement learning as an effective post-training strategy for MLLMs, but final-answer rewards remain sparse with respect to whether the reasoning process is actually grounded in visual evidence.

\subsection{Faithful Multimodal Reasoning}

Faithful multimodal reasoning requires both faithful perception of task-relevant visual evidence and faithful use of that evidence in the reasoning chain. Prior work shows that MLLMs frequently hallucinate visual content or rely on insufficient visual evidence~\citep{HallusionBench,liu2023mitigating,liu2024survey,lu2024relation}, and that longer reasoning chains may amplify hallucination when not properly grounded~\citep{More-Thinking,yao2025reasoning,lu2025piper}. Recent RL methods introduce perception- and grounding-aware signals: Perception-R1~\citep{PerceptionR1} rewards consistency with visual annotations, Vision-SR1~\citep{VisionSR1} decomposes responses into visual descriptions and language reasoning with self-rewarding, VPPO~\citep{VPPO} uses token-level visual dependency to reweight optimization toward visually pivotal tokens, and PAPO~\citep{wang2025perception} adds a perception-aware regularizer to the policy objective. A complementary line couples reasoning with explicit visual operations: Ground-R1~\citep{cao2025ground} rewards region selection, and DeepEyes~\citep{zheng2025deepeyes} learns to ``think with images'' by interleaving visual operations into the chain. Our work addresses both halves jointly: we strengthen perception through supervision native to the image and explicitly target the perception--reasoning disconnect, a failure mode that cuts across existing efforts in which correctly perceived evidence is dropped or contradicted during reasoning.

\section{Proposed Approach}
\label{sec:method}

\subsection{Overview}

\begin{figure}[!t]
  \centering
  \includegraphics[width=\linewidth]{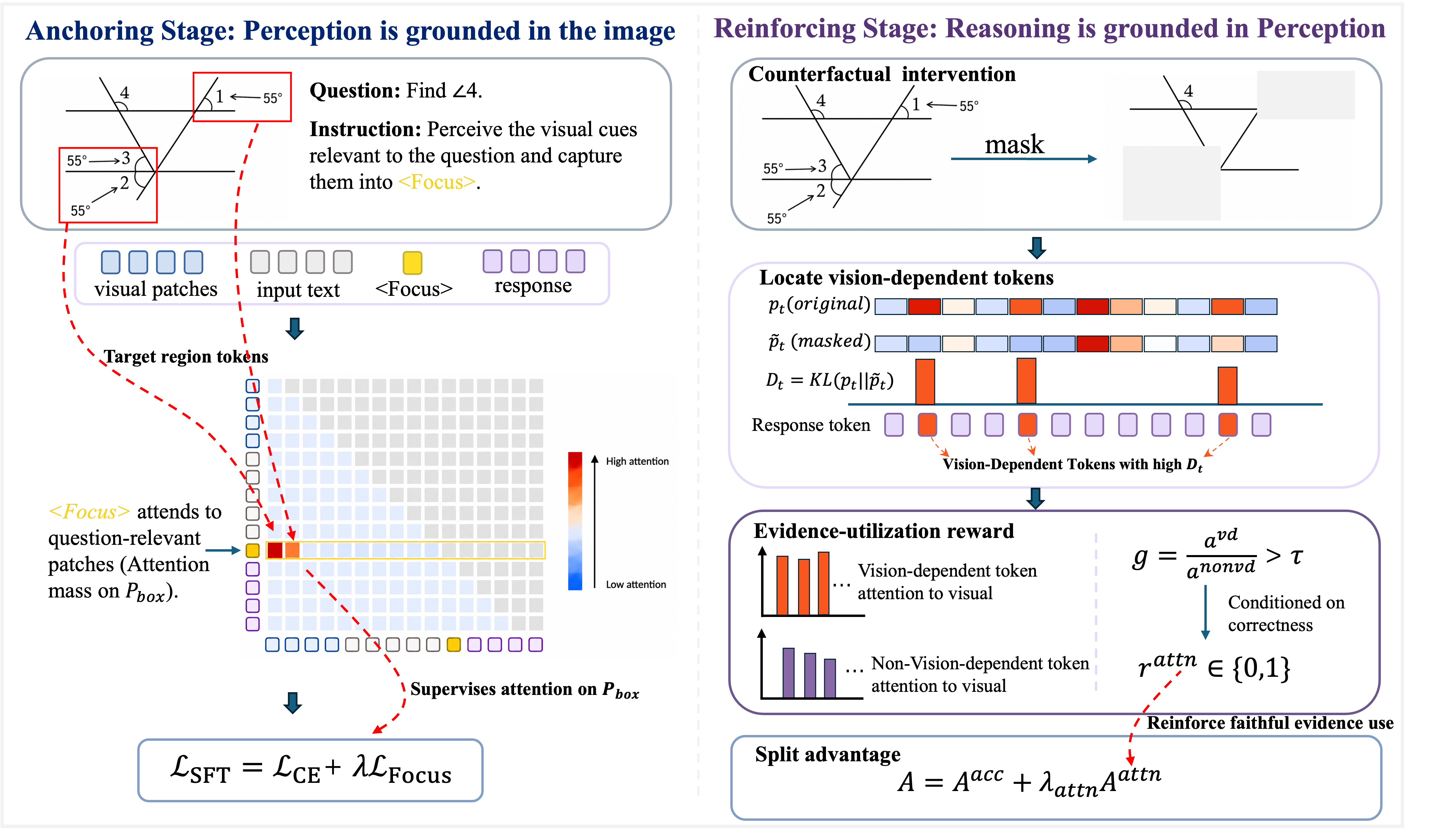}
  \caption{Overview of Faithful-MR1. \textbf{Left, Anchoring stage:} the \texttt{<Focus>} token's attention is supervised directly against the visual patch tokens covered by question-relevant bounding boxes (red boxes on the image and on the patch strip); the heatmap shows the supervised \texttt{<Focus>} attention row over visual patches. \textbf{Right, Reinforcing stage:} the policy is rolled out on both the original image and a counterfactually masked image; the per-token KL divergence between the two next-token distributions identifies high-$D_t$ (vision-dependent) tokens, and the ratio of their attention to visual content versus that of non-vision-dependent tokens is gated by answer correctness to form the evidence-utilization reward, which combines with the answer-correctness reward via the split GRPO advantage.}
  \label{fig:method_overview}
\end{figure}

Figure~\ref{fig:method_overview} summarizes Faithful-MR1. The Anchoring stage (Section~\ref{sec:anchor}) augments standard supervised fine-tuning with a region-grounded attention loss on a dedicated \texttt{<Focus>} token, anchoring perception to task-relevant image regions. The Reinforcing stage (Section~\ref{sec:reinforce}) initializes from the Anchoring-stage checkpoint and applies reinforcement learning: counterfactual image intervention localizes vision-dependent reasoning tokens, and a split GRPO advantage rewards answer-correct trajectories that concentrate visual attention at those tokens. 

\subsection{Anchoring Visual Attention}
\label{sec:anchor}

The Anchoring stage realizes ``focus first, then reason'' through a dedicated \texttt{<Focus>} token whose attention is supervised directly against task-relevant image regions.

\textbf{The \texttt{<Focus>} token.}
We prepend the instruction ``Perceive the visual cues relevant to the above question and capture them into \texttt{<Focus>}'' to the model's response, introducing \texttt{<Focus>} as a dedicated perceptual anchor for question-relevant visual evidence. \texttt{<Focus>} turns perception into an explicit one-step pre-reasoning subtask: before producing the full reasoning trajectory, the model first localizes and gathers the relevant evidence into \texttt{<Focus>}. Crucially, \texttt{<Focus>} is not a textual supervision target---it is an internal interface whose attention pattern, not its emitted tokens, carries the perceptual signal.

\textbf{Region-grounded attention supervision.}
For each sample, let $\mathcal{B}=\{b_1,\dots,b_M\}$ denote the bounding boxes of question-relevant evidence. Since the VLM consumes images as visual patch tokens, each box maps to a set $\mathcal{P}(b_m)$ of patch tokens; their union
\begin{equation}
\mathcal{P}_{\mathrm{box}} = \bigcup_{m=1}^{M} \mathcal{P}(b_m)
\end{equation}
defines the target region for \texttt{<Focus>}. Let $a^{(\ell,h)}_{\mathrm{Focus}}$ denote the attention row of \texttt{<Focus>} at layer $\ell$ and head $h$, and let
\begin{equation}
s^{(\ell,h)} = \sum_{i \in \mathcal{P}_{\mathrm{box}}} a^{(\ell,h)}_{\mathrm{Focus}, i}
\end{equation}
be its attention mass on target tokens. We supervise $s^{(\ell,h)}$ toward a target ratio $\tau$ via cross-entropy
\begin{equation}
\mathcal{L}^{(\ell,h)}_{\mathrm{focus}} = -\tau \log s^{(\ell,h)} - (1-\tau) \log\big(1 - s^{(\ell,h)}\big),
\end{equation}
averaged over a chosen set $\mathcal{S}$ of layer-head pairs:
\begin{equation}
\mathcal{L}_{\mathrm{Focus}} = \frac{1}{|\mathcal{S}|} \sum_{(\ell,h)\in \mathcal{S}} \mathcal{L}^{(\ell,h)}_{\mathrm{focus}}.
\end{equation}

\textbf{Hybrid objective.}
Combining the attention loss with the standard autoregressive answer loss
\begin{equation}
\mathcal{L}_{\mathrm{CE}} = -\sum_{t=1}^{T} \log p_{\theta}(y_t \mid I, q, y_{<t})
\end{equation}
gives the Anchoring objective
\begin{equation}
\mathcal{L}_{\mathrm{SFT}} = \mathcal{L}_{\mathrm{CE}} + \lambda \mathcal{L}_{\mathrm{Focus}},
\end{equation}
where $\lambda$ controls perceptual guidance strength. Answer supervision is preserved while \texttt{<Focus>}'s attention is trained to anchor perception to task-relevant image regions.

\subsection{Reinforcing Visual Attention}
\label{sec:reinforce}

The Reinforcing stage targets the perception--reasoning disconnect by directly optimizing faithful use---a property that, as Section~\ref{sec:intro} noted, is not directly observable in the reasoning trace. We expose it through counterfactual image intervention: response tokens whose predictions shift when the visual evidence is masked are treated as where vision causally matters, and answer-correct trajectories that concentrate visual attention at those tokens are rewarded. The policy initializes from the Anchoring-stage checkpoint and inherits its \texttt{<Focus>}-anchored prompt format, so that ``focus first'' behavior carries over into the RL phase.

\textbf{Counterfactual localization of vision-dependent tokens.}
We construct a perturbed image by masking the annotated regions while keeping the rest of the image unchanged. Given a sampled response, the policy performs one forward pass on the original image and another on the perturbed image. Let $p_t$ and $\tilde{p}_t$ denote the next-token distributions at response position $t$ under the original and perturbed images. The visual sensitivity of token $t$ is
\begin{equation}
D_t = D_{\mathrm{KL}}(p_t \,\|\, \tilde{p}_t).
\end{equation}
Tokens with high $D_t$ are causally vision-dependent: their predictions change when the critical visual evidence is removed. This turns evidence utilization from a trajectory-level notion into a token-level attribution problem---we explicitly localize the subset of response positions whose generation is causally supported by the image, in line with recent evidence that a small subset of \emph{critical tokens} disproportionately drives reasoning quality~\citep{lincritical}.

\textbf{Evidence-utilization reward.}
After identifying vision-dependent tokens, we reward trajectories that place stronger attention on visual content at those positions than at the remaining response tokens. Let $a^{\mathrm{vd}}$ be the average attention from vision-dependent reasoning tokens to visual tokens, and $a^{\mathrm{nonvd}}$ the corresponding average from non-vision-dependent response tokens. The evidence-utilization score
\begin{equation}
g = \frac{a^{\mathrm{vd}}}{a^{\mathrm{nonvd}} + \epsilon}
\end{equation}
is thresholded into a binary auxiliary reward $r^{\mathrm{attn}} = \mathbf{1}[g > \tau_g]$, with $\tau_g{=}2$ by default. This reward is gated on answer correctness: only answer-correct trajectories can receive it, ensuring that evidence-utilization credit is given to trajectories that both reach the right answer and visibly rely on visual evidence to do so.

\textbf{Split GRPO advantage.}
Let $r^{\mathrm{acc}} \in \{0,1\}$ be the binary answer-correctness reward. Rather than collapsing $r^{\mathrm{acc}}$ and $r^{\mathrm{attn}}$ before normalization, we compute GRPO~\citep{GRPO,schulman2017proximal} advantages separately,
\begin{equation}
A^{\mathrm{acc}} = \mathrm{GRPO}(r^{\mathrm{acc}}), \qquad
A^{\mathrm{attn}} = \mathrm{GRPO}(r^{\mathrm{attn}}),
\end{equation}
and combine them as $A = A^{\mathrm{acc}} + \lambda_{\mathrm{attn}} A^{\mathrm{attn}}$. The combined advantage drives a GRPO-style policy objective
\begin{equation}
\mathcal{L}_{\mathrm{RL}}(\theta)=
\mathbb{E}_{(I,q),o}\!\left[
\frac{1}{T}\sum_{t=1}^{T}
\min\!\big(
r_t(\theta)A,\,
\mathrm{clip}(r_t(\theta),1-\epsilon,1+\epsilon)A
\big)
\right]
- \beta\, D_{\mathrm{KL}}\!\left(\pi_{\theta}\,\|\,\pi_{\mathrm{ref}}\right),
\end{equation}
where $r_t(\theta)=\pi_{\theta}(o_t \mid I,q,o_{<t}) / \pi_{\theta_{\mathrm{old}}}(o_t \mid I,q,o_{<t})$ is the token-level importance ratio, $\epsilon$ is the clipping parameter, and $\beta$ controls the KL regularization to the reference policy. Answer correctness remains the primary signal. The attention reward additionally reinforces answer-correct trajectories that invoke visual evidence at exactly the reasoning steps where it causally matters.

\section{Experiments}

We evaluate our method on both mathematical reasoning and general multimodal understanding
benchmarks. We first describe the experimental setup, then present the main results.

\subsection{Experimental Setup}
\label{sec:exp}
\paragraph{Training data, models, and baselines.}
Our training data is constructed from the open-source Vision-SR1-47K dataset~\citep{VisionSR1}, with key visual regions annotated as bounding boxes; after filtering, we obtain 6K SFT examples and 13.2K RL examples (19.2K total). The bounding boxes supervise the \texttt{<Focus>} token in the Anchoring stage and define the masked regions for the Reinforcing stage's counterfactual image intervention. We use Qwen2.5-VL-Instruct~\citep{bai2025qwen25vl} as the backbone at the 3B and 7B scales. Under the same data setting we compare with GRPO~\citep{GRPO} and VPPO~\citep{VPPO}; we additionally compare with recent public multimodal reasoning checkpoints Vision-R1-7B~\citep{VisionR1}, Perception-R1-7B~\citep{PerceptionR1}, and Vision-SR1-7B~\citep{VisionSR1}, which are trained with substantially more data.

\paragraph{Benchmarks and metrics.}
We evaluate on seven benchmarks covering math reasoning (MathVision~\citep{MATHVision}, MathVerse~\citep{MathVerse}, MathVista~\citep{mathvista}, WeMath~\citep{WeMath}, DynaMath~\citep{DynaMath}) and general visual question answering (MMMU-Pro~\citep{MMMU-Pro}, HallusionBench~\citep{HallusionBench}). For DynaMath we report Worst Case Accuracy (WCA), the benchmark's headline robustness metric where a question is counted correct only if the model is correct on all 10 visual variants. We report per-benchmark accuracy and average the available scores as the overall result.

\paragraph{Implementation and evaluation.}
The Anchoring stage trains for one epoch with learning rate $1\times10^{-5}$; the auxiliary \texttt{<Focus>} attention loss uses empirically chosen weight $\lambda{=}0.1$ and target attention mass $\tau{=}0.8$ over the last 4 attention layers. The Reinforcing stage initializes from the Anchoring checkpoint and trains with 8 rollouts per prompt, rollout batch size 128, and learning rate $1\times10^{-6}$; the rule reward combines an answer-correctness reward and a format reward; the evidence-utilization reward uses threshold $\tau_g{=}2$ and coefficient $\lambda_{\mathrm{attn}}{=}0.3$ at the 7B scale and $0.1$ at the 3B scale. All evaluations use greedy decoding and are run with the EvalScope framework~\citep{evalscope_2024}; for reliability, we use Qwen3-235B-A22B-Instruct~\citep{qwen3technicalreport} as the LLM-as-a-judge across all benchmarks. Full data-construction and supplementary training details are in Appendix~\ref{app:additional_experimental_details}.

\begin{table}[!htbp]
  \centering
  \caption{Preliminary main results across multimodal benchmarks. Higher is better. Best results are shown in bold and second-best results are underlined. DynaMath is reported as Worst Case Accuracy (WCA), the benchmark's robustness metric, where a question is counted correct only if the model is correct on all 10 visual variants. ``--'' indicates the benchmark overlaps with the model's training set and is therefore invalid for comparison.}
  \label{tab:main_results}
  \footnotesize
  \setlength{\tabcolsep}{3pt}
  \renewcommand{\arraystretch}{1.15}
  \resizebox{\textwidth}{!}{%
  \begin{tabular}{llcccccccccc}
    \toprule
    \multirow{2}{*}{\textbf{Method}} & \multirow{2}{*}{\makecell[c]{\textbf{Training}\\\textbf{Data}}} & \multicolumn{6}{c}{\textbf{Math Reasoning}} & \multicolumn{3}{c}{\textbf{General Visual Question Answering}} & \multirow{2}{*}{\textbf{Overall}} \\
    \cmidrule(lr){3-8} \cmidrule(lr){9-11}
    & & \makecell[c]{\textbf{Math}\\\textbf{Vision}} & \makecell[c]{\textbf{Math}\\\textbf{Verse}} & \makecell[c]{\textbf{Math}\\\textbf{Vista}} & \textbf{WeMath} & \makecell[c]{\textbf{Dyna}\\\textbf{Math}} & \makecell[c]{\textbf{Math}\\\textbf{Avg}} & \textbf{MMMU-Pro} & \makecell[c]{\textbf{Hallusion}\\\textbf{Bench}} & \makecell[c]{\textbf{General}\\\textbf{Avg}} & \\
    \midrule
    \multicolumn{12}{l}{\textit{Public model comparison}} \\
    Vision-R1-7B & 210K & -- & 48.1 & 69.4 & -- & 24.6 & -- & 34.7 & 68.1 & 51.4 & -- \\
    Perception-R1-7B & 1.4K & 27.6 & 48.7 & \underline{73.2} & \underline{68.3} & \underline{25.2} & \underline{48.6} & 35.4 & \underline{70.0} & 52.7 & \underline{49.8} \\
    Vision-SR1-7B & 56K & \underline{28.3} & 47.8 & 69.8 & 64.2 & 23.4 & 46.7 & \textbf{41.1} & 68.1 & \underline{54.6} & 49.0 \\
    \midrule
    \multicolumn{12}{l}{\textit{Backbone model: Qwen2.5-VL-3B-Instruct}} \\
    Qwen2.5-VL-3B-Instruct & -- & 20.4 & 36.4 & 56.8 & 51.7 & 13.6 & 35.8 & 28.0 & 64.8 & 46.4 & 38.8 \\
    \quad + GRPO & 19.2K & 20.9 & 33.8 & 60.9 & 52.8 & 11.6 & 36.0 & \underline{30.2} & 67.2 & 48.7 & 39.6 \\
    \quad + VPPO & 19.2K & \underline{21.4} & \underline{36.7} & \underline{61.7} & \underline{53.8} & \underline{14.8} & \underline{37.7} & 29.8 & \underline{67.7} & 48.7 & \underline{40.8} \\
    \rowcolor{gray!15}
    \quad + \textbf{Faithful-MR1} & 19.2K & \textbf{23.6} & \textbf{40.9} & \textbf{61.9} & \textbf{62.8} & \textbf{18.6} & \textbf{41.6} & \textbf{31.7} & \textbf{68.0} & \textbf{49.9} & \textbf{43.9} \\
    \midrule
    \multicolumn{12}{l}{\textit{Backbone model: Qwen2.5-VL-7B-Instruct}} \\
    Qwen2.5-VL-7B-Instruct & -- & 23.8 & 45.6 & 68.8 & 61.0 & 20.4 & 43.9 & 34.3 & 65.0 & 49.7 & 45.5 \\
    \quad + GRPO & 19.2K & 26.5 & 48.7 & 70.9 & 60.9 & 22.2 & 45.8 & 36.6 & 69.3 & 53.0 & 47.9 \\
    \quad + VPPO & 19.2K & 27.1 & \underline{49.5} & 70.2 & 65.7 & \underline{25.2} & 47.5 & 37.8 & \textbf{70.8} & 54.3 & 49.5 \\
    \rowcolor{gray!15}
    \quad + \textbf{Faithful-MR1} & 19.2K & \textbf{28.3} & \textbf{51.9} & \textbf{73.5} & \textbf{68.9} & \textbf{26.8} & \textbf{49.9} & \underline{39.7} & 69.8 & \textbf{54.8} & \textbf{51.3} \\
    \bottomrule
  \end{tabular}}
\end{table}

\subsection{Main Results}

Table~\ref{tab:main_results} reports results across seven multimodal reasoning and perception benchmarks, organized into two complementary comparisons: the lower blocks compare training methods under matched Qwen2.5-VL-Instruct backbones~\citep{bai2025qwen25vl} and 19.2K training examples, while the upper block compares publicly released multimodal reasoning checkpoints trained at a range of data scales. Under matched backbone and data, Faithful-MR1 attains the best Overall score at both the 3B and 7B scales (43.9 and 51.3), and the lead extends across most individual benchmarks. Against publicly released checkpoints, Faithful-MR1 also surpasses Vision-SR1-7B and Vision-R1-7B on Overall and on most individual benchmarks despite using substantially less training data. This consistent advantage demonstrates the effectiveness of Faithful-MR1's two-stage design, which anchors perception to task-relevant image regions and reinforces the faithful use of that evidence in the reasoning chain.

Faithful-MR1's lead is most pronounced on math benchmarks, where the reasoning chain depends most heavily on faithful use of visual evidence; on General VQA, where tasks lean more on broad textual descriptions of the image, Faithful-MR1 still attains the best average. The same pattern carries over to the 3B scale. DynaMath, in particular, offers the strictest test of whether reasoning actually relies on the image: under its Worst Case Accuracy protocol, a question is counted correct only if the model answers all ten visual variants of the same problem correctly, so reasoning chains that bypass the image are penalized directly. Faithful-MR1 leads here, consistent with improvements arising from how visual evidence is propagated through the reasoning chain rather than from polishing perception text alone. 

\subsection{Ablation Study}
\label{sec:analysis}

To attribute Faithful-MR1's gains to the Anchoring and Reinforcing stages, Table~\ref{tab:two_stage_ablation} reports five configurations on the Qwen2.5-VL-7B-Instruct backbone, sharing the same training data, optimizer, and evaluation protocol as Table~\ref{tab:main_results}. The configurations form an attribution chain that exposes three independent deltas: row (3) over row (2) measures the effect of the Anchoring-stage data under standard SFT, separating it from the \texttt{<Focus>} mechanism; row (4) over row (3) isolates the \texttt{<Focus>} mechanism (the \texttt{<Focus>} token together with the auxiliary attention loss) on top of the same data; row (5) over row (4) isolates the Reinforcing stage's evidence-utilization reward initialized from the Anchoring-stage checkpoint.

Both stages contribute, with the Reinforcing stage driving the largest delta. Training on the Anchoring-stage data under standard SFT (row 3 over row 2) lifts Overall from 47.9 to 48.6, a modest data effect. The \texttt{<Focus>} mechanism adds a further 0.6 Overall points on the same data (row 4 over row 3, 48.6 to 49.2), confirming that supervising perception directly against image regions contributes beyond the data itself. The Reinforcing stage delivers the single largest delta (row 5 over row 4, 49.2 to 51.3, $+2.1$ Overall and $+2.6$ Math Avg), consistent with the finding that final accuracy is more directly influenced by faithful use than by faithful perception alone. Together, the two stages close the gap between vanilla GRPO and Faithful-MR1 ($+3.4$ Overall), with the bulk coming from the Reinforcing stage building on the perception anchored by the \texttt{<Focus>} mechanism.

\begin{table}[!htbp]
  \centering
  \caption{Ablation of Faithful-MR1's two-stage design on the Qwen2.5-VL-7B-Instruct backbone.
  \emph{Vanilla GRPO} denotes RL with only the answer-correctness reward;
  \emph{Vanilla SFT} denotes standard supervised fine-tuning on the Anchoring-stage data
  without the \texttt{<Focus>} token or the auxiliary attention loss;
  \emph{Anchoring} adds the \texttt{<Focus>} token and the auxiliary attention loss;
  \emph{Faithful-MR1} additionally enables the Reinforcing stage's evidence-utilization reward on top of the Anchoring-stage checkpoint.
  Best results in bold, second-best underlined.}
  \label{tab:two_stage_ablation}
  \footnotesize
  \setlength{\tabcolsep}{3pt}
  \renewcommand{\arraystretch}{1.15}
  \resizebox{\textwidth}{!}{%
  \begin{tabular}{llcccccccccc}
    \toprule
    \multirow{2}{*}{\textbf{\#}} & \multirow{2}{*}{\textbf{Configuration}} & \multicolumn{6}{c}{\textbf{Math Reasoning}} & \multicolumn{3}{c}{\textbf{General Visual Question Answering}} & \multirow{2}{*}{\textbf{Overall}} \\
    \cmidrule(lr){3-8} \cmidrule(lr){9-11}
    & & \makecell[c]{\textbf{Math}\\\textbf{Vision}} & \makecell[c]{\textbf{Math}\\\textbf{Verse}} & \makecell[c]{\textbf{Math}\\\textbf{Vista}} & \textbf{WeMath} & \makecell[c]{\textbf{Dyna}\\\textbf{Math}} & \makecell[c]{\textbf{Math}\\\textbf{Avg}} & \textbf{MMMU-Pro} & \makecell[c]{\textbf{Hallusion}\\\textbf{Bench}} & \makecell[c]{\textbf{General}\\\textbf{Avg}} & \\
    \midrule
    (1) & Qwen2.5-VL-7B-Instruct & 23.8 & 45.6 & 68.8 & 61.0 & 20.4 & 43.9 & 34.3 & 65.0 & 49.7 & 45.5 \\
    (2) & (1) + Vanilla GRPO & 26.5 & 48.7 & 70.9 & 60.9 & 22.2 & 45.8 & 36.6 & 69.3 & 53.0 & 47.9 \\
    (3) & \makecell[l]{(1) + Vanilla SFT + Vanilla GRPO} & 26.0 & \underline{50.0} & \underline{71.1} & 64.3 & 22.0 & 46.7 & 37.8 & 69.1 & 53.4 & 48.6 \\
    (4) & \makecell[l]{(1) + Anchoring + Vanilla GRPO} & \underline{27.5} & 49.5 & 70.9 & \underline{64.5} & \underline{24.2} & \underline{47.3} & \underline{38.2} & \underline{69.6} & \underline{53.9} & \underline{49.2} \\
    \rowcolor{gray!15}
    (5) & \makecell[l]{\textbf{Faithful-MR1}} & \textbf{28.3} & \textbf{51.9} & \textbf{73.5} & \textbf{68.9} & \textbf{26.8} & \textbf{49.9} & \textbf{39.7} & \textbf{69.8} & \textbf{54.8} & \textbf{51.3} \\
    \bottomrule
  \end{tabular}}
\end{table}

\begin{wrapfigure}[20]{r}{0.46\linewidth}
  \centering
  \includegraphics[width=\linewidth]{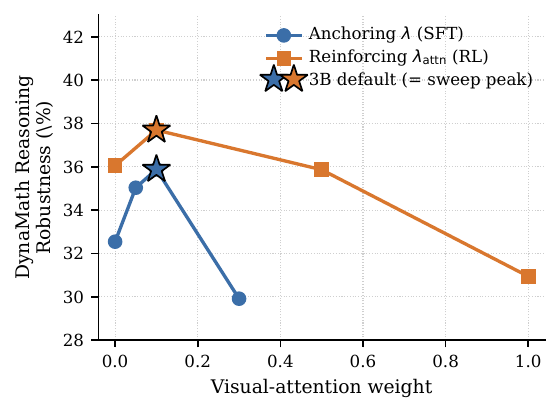}
  \caption{Effect of the Anchoring and Reinforcing weights on DynaMath Reasoning Robustness, sweeping the $\lambda$ and the $\lambda_{\mathrm{attn}}$ on Qwen2.5-VL-3B-Instruct. Stars mark the 3B defaults, set to the sweep peaks.}
  \label{fig:hyperparam_rr}
\end{wrapfigure}

\paragraph{Anchoring and Reinforcing weights: how much is enough?}
Each stage carries a single weight on its visual-attention term: $\lambda$ on the Anchoring focus loss in $\mathcal{L}_{\mathrm{SFT}}$, and $\lambda_{\mathrm{attn}}$ on the attention advantage in the Reinforcing split-GRPO objective. Both control the strength of visual-attention guidance during training. We sweep each independently on Qwen2.5-VL-3B-Instruct and report DynaMath's official Reasoning Robustness (Figure~\ref{fig:hyperparam_rr}), defined as the ratio $\mathrm{WCA}/\mathrm{AvgAcc}$ across the ten visual variants of each problem, which directly measures how consistently the model relies on the image. 
Both sweeps trace an inverted-U with peaks at moderate values: the Anchoring sweep peaks at $\lambda{=}0.1$ ($35.9$, $+3.3$ over $\lambda{=}0$), and the Reinforcing sweep at $\lambda_{\mathrm{attn}}{=}0.1$ ($37.7$, $+1.6$ over $\lambda_{\mathrm{attn}}{=}0$); pushing either weight too high ($\lambda{=}0.3$, $\lambda_{\mathrm{attn}}{=}1$) drops robustness below the corresponding zero-weight baseline. That both stages produce the same inverted-U is consistent with their acting on a shared visual-attention mechanism. We accordingly adopt the sweep optima $\lambda{=}0.1$ and $\lambda_{\mathrm{attn}}{=}0.1$ for our 3B main results.

\subsection{Quantifying the Perception-Reasoning Disconnect}
\label{sec:prd_analysis}

To separate gains in faithful perception from gains in faithful use, we evaluate behavioral signatures on the MathVerse Vision-Only split (788 responses), which removes figure-related text from the question and thus forces models to rely on the image. For each problem we pre-extract a list of \emph{gold visual facts} using Gemini-3-Flash conditioned on the image and the ground-truth answer; Gemini-3-Flash then scores every critical gold fact along two per-fact indicators: $P{=}1$ if the fact is verbalized correctly, and $U{=}1$ if it is used as a load-bearing premise in the derivation with a value consistent with the gold and not contradicted later in the chain. We report \emph{Perception Coverage}~$=\Pr(P{=}1)$, the rate at which gold facts are verbalized in the response; \emph{Faithful Use}~$=\Pr(U{=}1 \mid P{=}1)$, the rate at which verbalized facts are then committed to the derivation; and \emph{Answer Acc.}~$=\Pr(A{=}1)$. Perception Coverage and Faithful Use measure conceptually distinct quantities (a method can verbalize many visual facts decoratively, or commit fewer facts more faithfully to its derivation), so they need not move together.

Table~\ref{tab:disconnect_rate} shows that Perception Coverage and Accuracy do not move together. Vision-SR1-7B and Vision-R1-7B both attain high Perception Coverage but markedly lower Faithful Use; Vision-SR1 in particular leaves roughly half of its correctly verbalized facts unused or used with inconsistent values. Under matched backbone and training data, Faithful-MR1 reverses this profile, attaining the highest Faithful Use and Answer Accuracy most efficiently. The gain does not come from verbalizing more perception, but from committing a larger fraction of perceived evidence to the derivation, exactly the PRD reduction the Reinforcing stage targets.

\begin{table}[!htbp]
  \centering
  \caption{Behavioral analysis on the MathVerse Vision-Only split (788
  responses; percentages except Length). Perception Coverage~$=\Pr(P{=}1)$;
  Faithful Use~$=\Pr(U{=}1 \mid P{=}1)$;
  Length: mean tokens per response under the Qwen2.5-VL tokenizer;
  Answer Acc.\ $=\Pr(A{=}1)$. Best \textbf{bold},
  second-best \underline{underlined}.}
  \label{tab:disconnect_rate}
  \renewcommand{\arraystretch}{1.15}
  \begin{tabular}{lcccc}
    \toprule
    \textbf{Method} & \textbf{Perception Coverage} & \textbf{Faithful Use} & \textbf{Length (tok)} & \textbf{Answer Acc.} \\
    \midrule
    Vision-R1-7B           & \underline{70.2} & 54.3             & 526             & 41.4 \\
    Vision-SR1-7B          & \textbf{74.6}    & 49.4             & 597             & 41.8 \\
    \midrule
    Qwen2.5-VL-7B-Instruct & 65.2             & \underline{63.3} & 401             & 26.5 \\
    \quad + VPPO           & 65.1             & 62.9             & \underline{382} & \underline{41.9} \\
    \rowcolor{gray!15}
    \quad + \textbf{Faithful-MR1}  & 61.3             & \textbf{66.7}    & \textbf{325}    & \textbf{45.1} \\
    \bottomrule
  \end{tabular}
\end{table}

\subsection{Qualitative Visualization}

\begin{figure}[t]
  \centering
  \includegraphics[width=\linewidth]{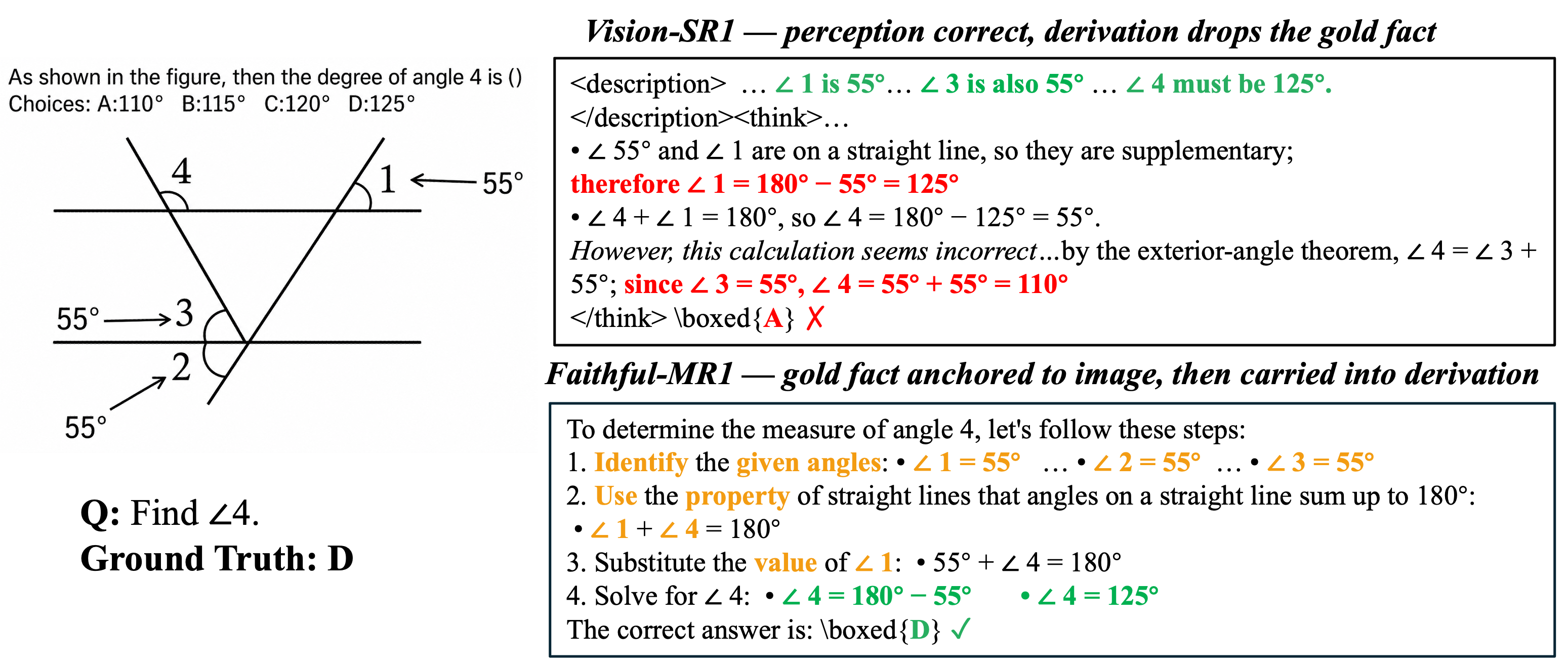}
  \caption{Case-level illustration of the Faithful Use gap. \textbf{Left:} problem image, with gold facts $\angle 1{=}55^\circ$ and $\angle 3{=}55^\circ$. \textbf{Top-right:} Vision-SR1's response; the derivation re-assigns $125^\circ$ to $\angle 1$ (red span). \textbf{Bottom-right:} Faithful-MR1's response; tokens identified as vision-dependent (high visual-sensitivity score $D_t$) are shaded in orange, and the load-bearing arithmetic step is highlighted in green.}
  \label{fig:qualitative_visualization}
\end{figure}

To make the Faithful Use gap of Table~\ref{tab:disconnect_rate} concrete,
Figure~\ref{fig:qualitative_visualization} contrasts Vision-SR1 and Faithful-MR1
on the same MathVerse Vision-Only problem.
Both models verbalize the gold facts for solving the problem ($\angle 1{=}55^\circ$, $\angle 3{=}55^\circ$),
yet Vision-SR1 silently re-assigns $125^\circ$ to $\angle 1$ inside its derivation
and never recovers, a textbook PRD failure. In contrast, Faithful-MR1's high-$D_t$
tokens (those most sensitive to image masking) concentrate on the gold-value tokens
themselves, and $\angle 1{=}55^\circ$ is then carried faithfully into the load-bearing
arithmetic step that yields the correct answer, exposing where vision causally matters
and how it propagates through the chain.

\section{Limitations}

While Faithful-MR1 delivers consistent gains across multimodal reasoning benchmarks on both the 3B and 7B variants of Qwen2.5-VL with substantially less training data, there are two limitations including its reliance on region-level annotations and the additional training-time compute introduced by counterfactual intervention. The Anchoring stage requires bounding-box annotations of question-relevant regions, which we obtain at scale from an automated multimodal annotator (Gemini-3-Flash). In addition, the Reinforcing stage's counterfactual image intervention runs one extra masked-image forward pass per rollout during training, a modest overhead over vanilla GRPO; inference itself adds no overhead, and Faithful-MR1 produces shorter, more token-efficient responses than recent baselines (Table~\ref{tab:disconnect_rate}).

\section{Conclusion}

Multimodal RLVR yields smaller gains than its text-only counterpart: faithful multimodal reasoning requires both faithful perception and faithful use of task-relevant visual evidence, but current multimodal RLVR methods address these only partially. Specifically, perception is typically supervised through generated text and faithful use is often overlooked, exposing the perception--reasoning disconnect where correctly perceived evidence is dropped or contradicted during reasoning. Our quantitative analysis confirms PRD as a distinct failure mode beyond the reach of perception supervision alone, with final accuracy more directly influenced by faithful use than by faithful perception alone. To close these gaps, we proposed {Faithful-MR1}, a training framework that anchors and reinforces visual attention: the {Anchoring stage} supervises a \texttt{<Focus>} token's attention directly against image regions rather than generated text, while the {Reinforcing stage} rewards answer-correct trajectories that concentrate visual attention where vision causally matters, exposed through counterfactual image intervention. With substantially less training data, Faithful-MR1 outperforms recent multimodal reasoning baselines on both Qwen2.5-VL-Instruct 3B and 7B backbones.

\bibliographystyle{plainnat}
\bibliography{references}

\appendix

\section{Additional Experimental Details}
\label{app:additional_experimental_details}

\subsection{Training and Evaluation Protocol}

\paragraph{Training data construction.}
We construct our training data from the open-source Vision-SR1-47K dataset~\citep{VisionSR1}. Key visual regions are annotated as bounding boxes by Gemini-3-Flash~\citep{Gemini3Flash} conditioned on the image, query, and ground-truth answer (full prompt in Appendix~\ref{app:prompts}). For the SFT split, responses are generated by rollouts from Qwen2.5-VL-7B-Instruct~\citep{bai2025qwen25vl} and filtered by rules and an LLM-as-a-judge for answer correctness and reasoning quality. After filtering we obtain 6K SFT examples and 13.2K RL examples (19.2K total). All matched-backbone baselines use this same split; GRPO uses only the final-answer reward, while VPPO and our method use the same answer reward with their respective faithfulness training signals.

\paragraph{Implementation details.}
We implement the Anchoring stage with LlamaFactory~\citep{zheng2024llamafactory} and the Reinforcing stage with EasyR1~\citep{zheng2025easyr1,sheng2024hybridflow}, both with bf16 precision; rollout temperature is set to 1.0 in the Reinforcing stage. GRPO and VPPO baselines are reproduced from their official open-source implementations on our training data. All training runs are conducted on 32 AMD MI308X GPUs under the ROCm/HIP runtime.

\subsection{Additional Case Studies}
\label{app:additional_cases}

To complement the qualitative example in Figure~\ref{fig:qualitative_visualization}, we provide three additional MathVerse Vision-Only cases where a baseline correctly verbalizes the gold visual fact ($P{=}1$) but fails to use it faithfully in the derivation ($U{=}0$, $A{=}0$), while Faithful-MR1 reaches the correct answer through faithful use of the same fact. We show each model's full response with three highlights: \textcolor{blue!75!black}{blue} marks where the gold visual fact is correctly verbalized, \textcolor{red!75!black}{red} marks where the baseline drops or contradicts that fact in the derivation (the perception--reasoning disconnect), and \textcolor{green!50!black}{green} marks the load-bearing step in Faithful-MR1's response that uses the fact correctly. The three cases involve different baselines (Vision-SR1, VPPO, Perception-R1), illustrating that the perception--reasoning disconnect spans different methods rather than being specific to one design.

\paragraph{Case A: angle bisector with parallel lines (Vision-SR1).}

\begin{figure}[!htbp]
  \centering
  \includegraphics[width=0.4\linewidth]{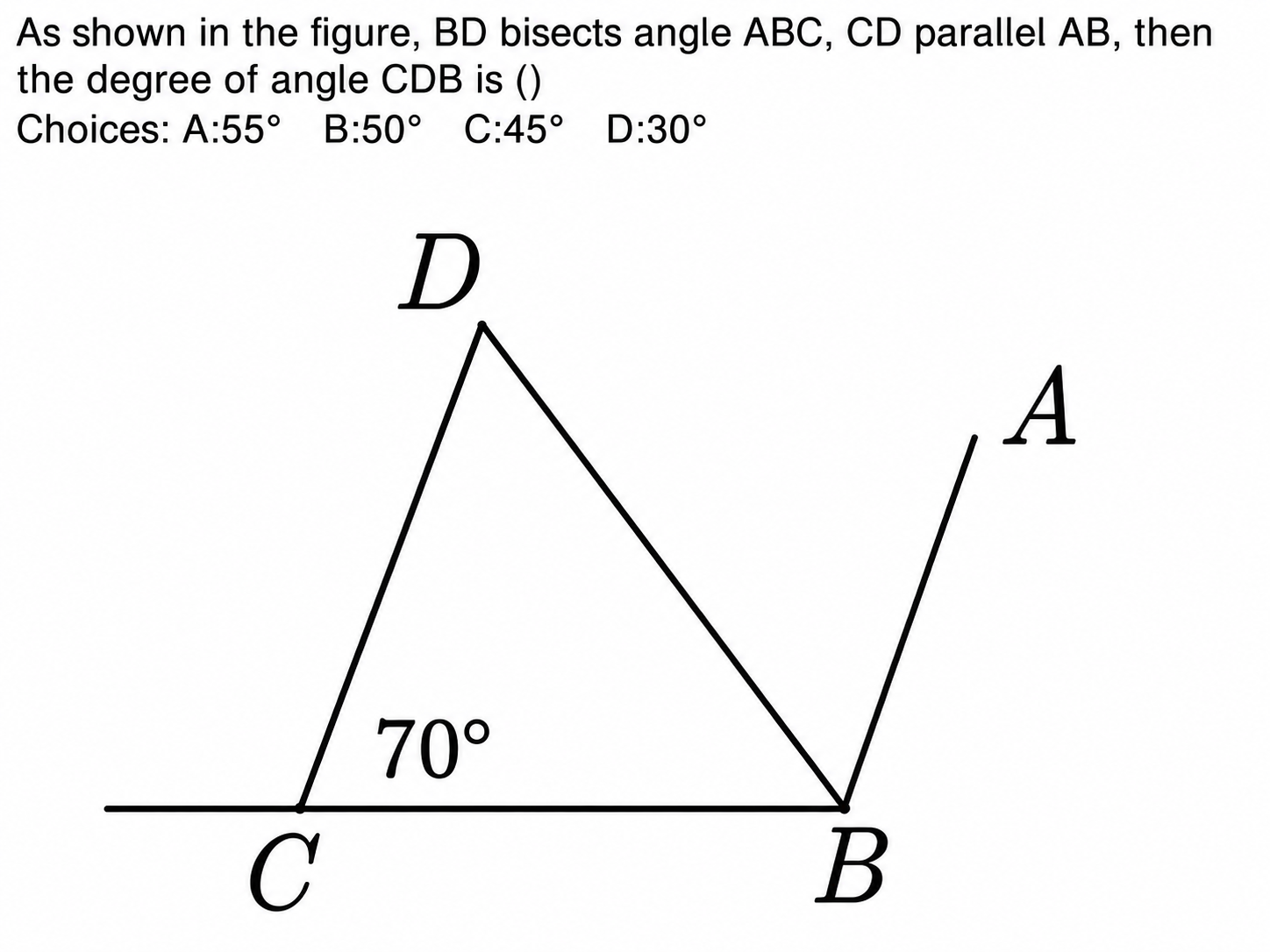}
\end{figure}

\noindent\textbf{Question.} $BD$ bisects $\angle ABC$, $CD \parallel AB$; find $\angle CDB$. Choices: A:$55^\circ$\quad B:$50^\circ$\quad C:$45^\circ$\quad D:$30^\circ$. \textbf{GT: A ($55^\circ$).}

\noindent\textbf{Gold visual fact.} $\angle DCB$ is labeled $70^\circ$ in the figure.

\medskip
\noindent\textbf{Vision-SR1 response (PRD; $P{=}1$, $U{=}0$, $A{=}0$):}
\begin{promptbox}
\footnotesize
\texttt{<description>}\\
The figure shows a triangle $ABC$ with point $D$ on the extension of side $CB$ beyond $B$. Segment $BD$ bisects angle $ABC$, and line segment $CD$ is parallel to line segment $AB$. \textcolor{blue!75!black}{The angle at $C$ is labeled $70^\circ$.} Because $CD$ is parallel to $AB$ and $BD$ bisects angle $ABC$, \textcolor{red!75!black}{angle $CDB$ is equal to half of angle $ACB$ (which is $70^\circ$), making it $35^\circ$.} However, the problem asks for the measure of angle $CDB$, and the closest match among the given choices is $30^\circ$.\\
\texttt{</description>}

\medskip
\texttt{<think>}\\
To solve this problem, we need to analyze the geometric properties and relationships in the given figure.

\textbf{1. Understanding the Figure:}
We have a triangle $\triangle ABC$ with point $D$ on the extension of side $CB$. $BD$ bisects angle $\angle ABC$. $CD$ is parallel to $AB$. We need to find the measure of angle $\angle CDB$.

\textbf{2. Angle Bisector Property:}
Since $BD$ bisects $\angle ABC$, it divides $\angle ABC$ into two equal angles. Let $\angle ABD = \angle DBC = x$.

\textbf{3. Using Parallel Lines:}
$CD \parallel AB$. By alternate interior angles, $\angle DBC = \angle BAC = x$.

\textbf{4. Triangle Properties:}
In triangle $BCD$, angles $\angle DBC$ and $\angle BDC$ are equal because $\triangle BCD$ is isosceles (since $CD \parallel AB$ and angles are equal). Therefore, $\angle CDB = x$.

\textbf{5. Finding the Angle:}
Since $\angle ABD = x$ and $\angle DBC = x$, and knowing that $\angle ABC = 2x$. \textcolor{red!75!black}{The measure of angle $\angle CDB$ is half of the angle $\angle ACB$ because of the parallel lines and angle bisector. Given that $\angle ACB = 70^\circ$, angle $\angle CDB = \frac{70^\circ}{2} = 35^\circ$.}

\textbf{6. Matching with Options:}
The closest option to $35^\circ$ is not listed, so we must re-evaluate the problem constraints or options. However, given the problem constraints and options, it seems possible there might be a mistake in calculation or assumption. Let's consider the closest feasible option, which might be due to rounding or problem setup error.

Upon reviewing the problem and options, the closest feasible option to our calculated $35^\circ$ is not available. However, given the problem setup, it seems likely the intended answer might be one of the listed options.

Upon re-evaluation, the closest option to $35^\circ$ is $30^\circ$, but given the problem constraints, it seems likely the intended answer might be $45^\circ$ or $50^\circ$ due to possible rounding or error.

\textit{[The model then enters a loop, repeating the paragraph ``Upon final verification, the closest option to our calculated $35^\circ$ is not available, but given the problem constraints, it seems likely the intended answer might be due to error or setup.'' approximately 25 times before truncation. Final answer: $30^\circ$ (option D, incorrect).]}\\
\texttt{</think>}
\end{promptbox}

\medskip
\noindent\textbf{Faithful-MR1 response ($P{=}1$, $U{=}1$, $A{=}1$):}
\begin{promptbox}
\footnotesize
To determine the measure of angle $\angle CDB$:

\textbf{1. Identify the given information:} $BD$ bisects $\angle ABC$; $CD \parallel AB$; \textcolor{blue!75!black}{$\angle DCB = 70^\circ$}.

\textbf{2. Use the property of angle bisector:} Since $BD$ bisects $\angle ABC$, we have $\angle ABD = \angle DBC$.

\textbf{3. Use the property of parallel lines:} Since $CD \parallel AB$, the corresponding angles are equal: $\angle ABD = \angle BDC$.

\textbf{4. Relate the angles using the straight line property:} The angles on a straight line sum up to $180^\circ$:
\textcolor{green!50!black}{$\angle DCB + \angle BDC + \angle DBC = 180^\circ$.}
Substitute \textcolor{blue!75!black}{$\angle DCB = 70^\circ$} and $\angle BDC = \angle ABD$:
\textcolor{green!50!black}{$70^\circ + \angle ABD + \angle ABD = 180^\circ$,} so $2\angle ABD = 110^\circ$ and \textcolor{green!50!black}{$\angle ABD = 55^\circ$.}

\textbf{5. Conclusion:} Since $\angle ABD = \angle DBC$, $\angle DBC = 55^\circ$. The measure of $\angle CDB$ is $\boxed{A}$.
\end{promptbox}

\paragraph{Case B: cone height from a sector (VPPO).}

\begin{figure}[!htbp]
  \centering
  \includegraphics[width=0.4\linewidth]{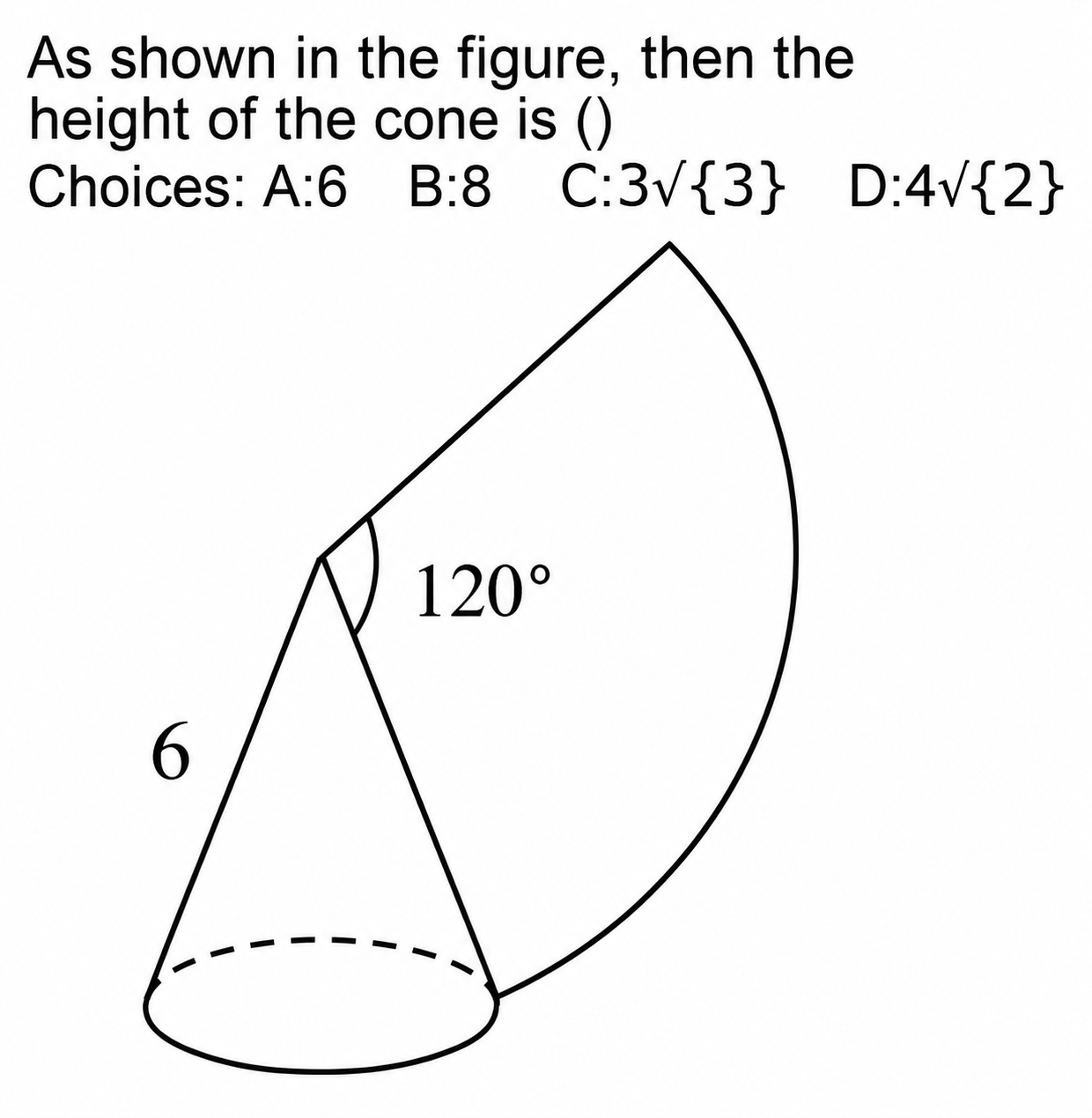}
\end{figure}

\noindent\textbf{Question.} A cone is formed by rolling up a sector with slant height $6$ and central angle $120^\circ$; find the height of the cone. Choices: A:$6$\quad B:$8$\quad C:$3\sqrt{3}$\quad D:$4\sqrt{2}$. \textbf{GT: D ($4\sqrt{2}$).}

\noindent\textbf{Gold visual facts.} (F1) The slant height of the cone is labeled $6$. (F2) The central angle of the sector forming the lateral surface is labeled $120^\circ$.

\medskip
\noindent\textbf{VPPO response (PRD; $P{=}1$, $U{=}0$, $A{=}0$):}
\begin{promptbox}
\footnotesize
To determine the height of the cone, we need to analyze the given information and use geometric properties. Here's the step-by-step reasoning:

\textbf{1. Identify the given information:} \textcolor{blue!75!black}{The slant height of the cone is $6$.} \textcolor{blue!75!black}{The angle at the vertex of the cone is $120$ degrees.}

\textbf{2. Understand the geometry:} \textcolor{red!75!black}{The angle at the vertex of the cone is $120$ degrees, which means the angle at the base of the cone is $60$ degrees (since the angle at the vertex is split into two equal angles of $60$ degrees each).} The slant height of the cone is the hypotenuse of a right triangle formed by the height of the cone, the radius of the base, and the slant height.

\textbf{3. Use trigonometry:} In the right triangle formed by the height of the cone, the radius of the base, and the slant height, the angle at the base is $60$ degrees. The height of the cone is the opposite side to the $60$-degree angle, and the slant height is the hypotenuse. The relationship in a $30$-$60$-$90$ triangle tells us that the height (opposite the $60$-degree angle) is $\frac{\sqrt{3}}{2}$ times the hypotenuse. Therefore, \textcolor{red!75!black}{the height $h$ of the cone is $6 \times \frac{\sqrt{3}}{2} = 3\sqrt{3}$.}

So, the height of the cone is $3\sqrt{3}$.

The correct answer is $\boxed{C}$.
\end{promptbox}

\medskip
\noindent\textbf{Faithful-MR1 response ($P{=}1$, $U{=}1$, $A{=}1$):}
\begin{promptbox}
\footnotesize
To find the height of the cone, let's analyze the given information:

\textbf{1.} \textcolor{blue!75!black}{The slant height of the cone is $6$ units.} \textcolor{blue!75!black}{The central angle of the sector is $120^\circ$.}

The sector of the cone forms a part of a circle with the radius equal to the slant height of the cone. The arc length of this sector is equal to the circumference of the base of the cone.

The formula for the arc length $L$ of a sector with radius $r$ and central angle $\theta$ (in degrees) is:
\[L = \frac{\theta}{360^\circ} \times 2\pi r.\]

Here, $L = 2\pi r$ (circumference of the base of the cone) and the sector radius is $6$:
\textcolor{green!50!black}{\[2\pi r = \frac{120^\circ}{360^\circ} \times 2\pi \times 6 = \frac{1}{3} \times 12\pi = 4\pi,\]
so $r = 2$.}

Now, we have the radius of the base of the cone as $2$ units. To find the height $h$ of the cone, we use the Pythagorean theorem in the right triangle formed by the height, the radius of the base, and the slant height:
\[h^2 + r^2 = l^2 \implies h^2 + 2^2 = 6^2 \implies h^2 = 32 \implies \textcolor{green!50!black}{h = \sqrt{32} = 4\sqrt{2}}.\]

Thus, the height of the cone is $\boxed{D}$.
\end{promptbox}

\paragraph{Case C: tangents from an external point (Perception-R1).}

\begin{figure}[!htbp]
  \centering
  \includegraphics[width=0.4\linewidth]{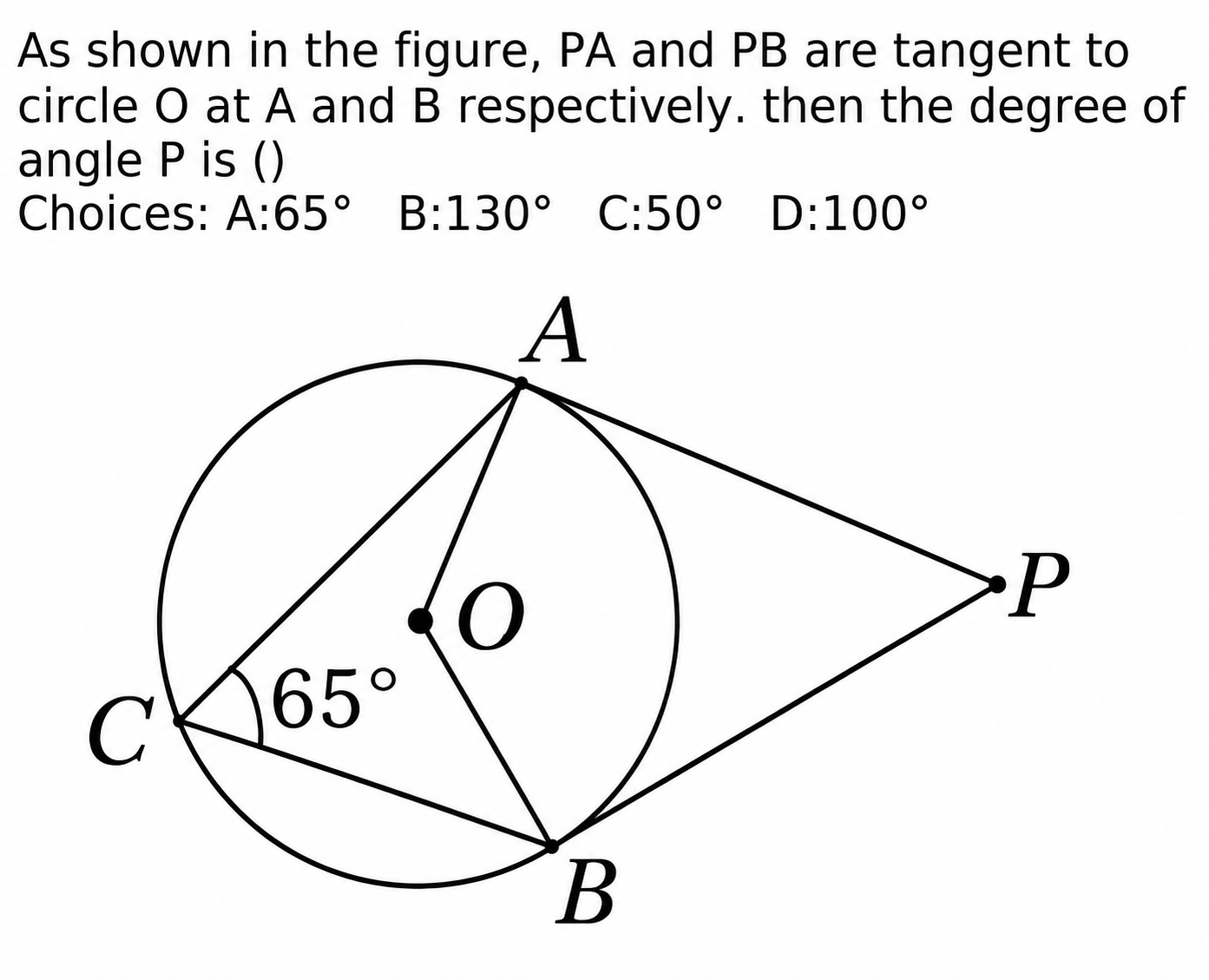}
\end{figure}

\noindent\textbf{Question.} $PA$ and $PB$ are tangent to circle $O$ at $A$ and $B$; find $\angle P$. Choices: A:$65^\circ$\quad B:$130^\circ$\quad C:$50^\circ$\quad D:$100^\circ$. \textbf{GT: C ($50^\circ$).}

\noindent\textbf{Gold visual fact.} $\angle ACB$ is labeled $65^\circ$ in the figure.

\medskip
\noindent\textbf{Perception-R1 response (PRD; $P{=}1$, $U{=}0$, $A{=}0$):}
\begin{promptbox}
\footnotesize
Looking at the figure, we know that $PA$ and $PB$ are tangents to the circle at points $A$ and $B$ respectively. By the tangent--tangent theorem, the tangents drawn from an external point to a circle are equal in length, and the angle between a tangent and a radius at the point of tangency is $90$ degrees. Therefore, $\angle OAP = \angle OBP = 90^\circ$.

Since \textcolor{blue!75!black}{$\angle ACB = 65^\circ$}, we can use the fact that $\angle ACB$ is an angle subtended by arc $AB$ at point $C$ on the circle. The angle $\angle AOB$ subtended by the same arc at the center is twice the angle subtended at any point on the circle, so $\angle AOB = 2 \times 65^\circ = 130^\circ$.

Now, consider $\triangle OAP$ and $\triangle OBP$. Both are right triangles with $\angle OAP = \angle OBP = 90^\circ$. The line $OP$ is a straight line through the center $O$, so $\angle AOB = 130^\circ$. The angle $\angle APB$ can be found using the fact that the sum of the angles in $\triangle AOB$ is $180^\circ$, and also using the external angle property of the tangents.

Since $PA$ and $PB$ are tangents, $\angle OAP = \angle OBP = 90^\circ$, and the arc $\widehat{AB}$ subtends $\angle AOB = 130^\circ$. The angle $\angle APB$ is the external angle to $\angle AOB$ with respect to the tangents, and it can also be found using the fact that the sum of the angles around point $P$ is $360^\circ$ and the angles $\angle OAP$ and $\angle OBP$ are $90^\circ$ each.

Using the fact that $\angle AOB = 130^\circ$, the angle $\angle APB$ can be found as the supplementary angle to the arc subtended by $\angle ACB$ because \textcolor{red!75!black}{$\angle APB = 360^\circ - \angle OAP - \angle OBP - \angle AOB = 360^\circ - 90^\circ - 90^\circ - 130^\circ = 140^\circ - 130^\circ = 50^\circ$} because $\angle APB = 180^\circ - \angle ACB = 180^\circ - 65^\circ = 115^\circ$ \textcolor{red!75!black}{but considering the supplementary angle directly is $\boxed{130}$.}

\texttt{<answer>}\\
$\boxed{130}$\\
\texttt{</answer>}
\end{promptbox}

\medskip
\noindent\textbf{Faithful-MR1 response ($P{=}1$, $U{=}1$, $A{=}1$):}
\begin{promptbox}
\footnotesize
To determine the measure of angle $P$:

\textbf{1. Identify the given information:} $PA$ and $PB$ are tangents to the circle at points $A$ and $B$ respectively; the measure of angle \textcolor{blue!75!black}{$\angle ACB = 65^\circ$}.

\textbf{2. Use the property of tangents:} The angle between a tangent and a radius at the point of tangency is $90^\circ$. Therefore, $\angle OAP = 90^\circ$ and $\angle OBP = 90^\circ$.

\textbf{3. Consider the quadrilateral $OAPB$:} The sum of the interior angles of a quadrilateral is $360^\circ$. The angles $\angle OAP$ and $\angle OBP$ are each $90^\circ$.

\textbf{4. Write the equation for the sum of the angles in quadrilateral $OAPB$:} $\angle OAP + \angle OBP + \angle AOB + \angle P = 360^\circ$, giving $180^\circ + \angle AOB + \angle P = 360^\circ$.

\textbf{5. Use the property of the central angle:} The central angle $\angle AOB$ is twice the inscribed angle $\angle ACB$:
\textcolor{green!50!black}{$\angle AOB = 2 \times \angle ACB = 2 \times 65^\circ = 130^\circ.$}

\textbf{6. Substitute back into the equation:} $180^\circ + 130^\circ + \angle P = 360^\circ$, so \textcolor{green!50!black}{$\angle P = 360^\circ - 310^\circ = 50^\circ$.}

The measure of angle $P$ is $\boxed{C}$.
\end{promptbox}

\subsection{Prompts}
\label{app:prompts}

This section lists the prompts used at every stage where Faithful-MR1 invokes an LLM or a VLM as a pipeline component. Placeholders are written as \texttt{\{NAME\}} and are filled in at call time.

\paragraph{(P1) Bounding-box region annotation (Gemini-3-Flash).}
Conditioned on the image, query, and ground-truth answer, we ask Gemini-3-Flash to localize the visual entities required to verify the answer as bounding boxes (normalized to $[0, 1000]$).

\begin{promptbox}
\footnotesize
\textbf{Role}: You are a Lead Visual Grounding Engineer specializing in Evidence-Based Localization. Your goal is to identify and ground only the essential visual anchors that constitute direct proof for the \texttt{\{QUERY\}} and \texttt{\{ANSWER\}}.

\smallskip
\textbf{Task Objective}: Analyze the \texttt{\{IMAGE\}}, \texttt{\{QUERY\}}, and \texttt{\{ANSWER\}}. Your mission is to provide precise bounding boxes for the essential visual entities strictly required to verify the correctness of the \texttt{\{ANSWER\}}. Do not label background clutter or entities that do not directly support the logical derivation from the \texttt{\{QUERY\}} to the \texttt{\{ANSWER\}}.

\smallskip
\textbf{Strict Annotation Principles}:
\begin{enumerate}[leftmargin=1.6em, labelsep=0.4em, itemsep=2pt, topsep=2pt, parsep=0pt]
  \item \textbf{Evidence-Centricity (Primary Rule)}: Only annotate objects that act as ``smoking guns'' for the answer.
  \item \textbf{Atomic Localization}: Label individual, specific objects. Avoid ``scene-level'' or ``group'' boxes.
  \item \textbf{Tightness}: Boxes should closely fit the visible object boundaries with reasonable margins that respect the object's natural contours. Avoid excessive padding, but allow slight breathing room to preserve the object's context and readability.
  \item \textbf{Exclusion of Contextual Noise}: Do not annotate:
    \begin{itemize}[leftmargin=1.2em, labelsep=0.4em, itemsep=1pt, topsep=2pt, parsep=0pt]
      \item Large environmental or structural elements (e.g., ``ground'', ``wall'', ``sky'', ``full figure'', or ``full table'') unless they are the direct subject of the query.
      \item Implied objects that are not clearly visible.
      \item Objects mentioned in the answer for purely descriptive/decorative purposes that don't assist in the logical deduction.
    \end{itemize}
  \item \textbf{Vision-Only Scenarios}: In cases where the query and options are rendered directly within the \texttt{\{IMAGE\}}, you must explicitly provide bounding boxes for the query text and every individual option, regardless of which one is the correct \texttt{\{ANSWER\}}.
\end{enumerate}

\smallskip
\textbf{Reasoning Process (Chain-of-Thought)}: Before generating coordinates, perform the following mental steps:
\begin{enumerate}[leftmargin=1.6em, labelsep=0.4em, itemsep=2pt, topsep=2pt, parsep=0pt]
  \item \textbf{Core Subject Identification}: What are the primary subjects in the \texttt{\{IMAGE\}} required to justify the \texttt{\{ANSWER\}} for the \texttt{\{QUERY\}}?
  \item \textbf{Causal Filter}: Does this specific entity provide direct supporting or refuting evidence to justify the \texttt{\{ANSWER\}} for the \texttt{\{QUERY\}}? (e.g., If the query asks if a car is ``red'' and the answer is ``Incorrect'', the car itself is the refuting evidence to show its actual color, while the road it sits on is irrelevant.)
  \item \textbf{Spatial Mapping}: Precisely locate the regions of the remaining ``Essential Anchors.''
\end{enumerate}

\smallskip
\textbf{Output Format}: Return a valid JSON object only. Use normalized coordinates \texttt{[ymin, xmin, ymax, xmax]} in the range of \texttt{[0, 1000]}.

\begin{Verbatim}
{
  "reasoning_process": "Provide a concise step-by-step analysis following the 3-step Reasoning Process: 1) Core Subject Identification, 2) Causal Filter, and 3) Spatial Mapping.",
  "objects": [
    {
      "name": "specific_object_name",
      "box_2d": [ymin, xmin, ymax, xmax]
    }
  ]
}
\end{Verbatim}
\end{promptbox}

\paragraph{(P2) Training and inference prompt template.}
The shared prompt template used by our method for both training stages and evaluation inference.

\begin{promptbox}
\begin{Verbatim}
<image>

{PROBLEM}

Perceive the visual cues relevant to the above question and capture them into <Focus>.

Think step by step, then provide your final answer enclosed within `\boxed{}` at the end of your response. For multiple-choice questions, place only the option letter (e.g., A) in `\boxed{}`. For other questions, place the final numerical or text answer in `\boxed{}`.
\end{Verbatim}
\end{promptbox}

\paragraph{(P3) LLM-as-a-judge (Qwen3-235B-A22B-Instruct).}
We use Qwen3-235B-A22B-Instruct via EvalScope to compare each model's predicted answer against the ground truth and return a single letter \texttt{A} (correct) or \texttt{B} (incorrect).

\begin{promptbox}
\begin{Verbatim}
Your job is to look at a question, a gold target, and a predicted answer, and return a letter "A" or "B" to indicate whether the predicted answer is correct or incorrect.

[Question]
{QUESTION}

[Reference Answer]
{GOLD}

[Predicted Answer]
{PRED}

Evaluate the model's answer based on correctness compared to the reference answer.
Grade the predicted answer of this new question as one of:
A: CORRECT
B: INCORRECT

Just return the letters "A" or "B", with no text around it.
\end{Verbatim}
\end{promptbox}

\paragraph{(P4) Gold visual facts extraction.}
Conditioned on the image and ground-truth answer, we ask Gemini-3-Flash to enumerate the minimal set of visual facts on which any correct solution must rely.

\begin{promptbox}
\footnotesize
You are extracting the visual facts required to solve a multimodal math problem.

\smallskip
\textbf{Question}\\
\texttt{\{PROBLEM\}}

\smallskip
\textbf{Ground-truth answer}\\
\texttt{\{GROUND\_TRUTH\}}

\smallskip
\textbf{Task}: Identify the \emph{minimal set} of visual facts (numeric labels, geometric relations, spatial configurations, table entries, chart values, \ldots) that any correct solution must rely on.

\smallskip
For each fact output:
\begin{itemize}[leftmargin=1.2em, labelsep=0.4em, itemsep=1pt, topsep=2pt, parsep=0pt]
  \item \texttt{"id"}: short unique identifier (\texttt{"F1"}, \texttt{"F2"}, \ldots)
  \item \texttt{"fact"}: one canonical sentence stating the fact
  \item \texttt{"criticality"}: \texttt{"critical"} if the answer is unreachable without it, else \texttt{"supporting"}
  \item \texttt{"fact\_type"}: one of \texttt{"numeric\_label"}, \texttt{"geometric\_relation"}, \texttt{"spatial\_position"}, \texttt{"object\_count"}, \texttt{"chart\_value"}, \texttt{"table\_entry"}, \texttt{"other"}
\end{itemize}

\smallskip
\textbf{Rules}:
\begin{enumerate}[leftmargin=1.6em, labelsep=0.4em, itemsep=2pt, topsep=2pt, parsep=0pt]
  \item List ONLY facts that the IMAGE provides; do not list facts that are already stated in the question text.
  \item Be conservative: if a fact is implied or redundant, mark it \texttt{"supporting"}.
  \item Use the ground-truth answer ONLY to decide which image facts are needed; never copy the answer or any derivation conclusion into the fact list.
  \item Aim for 1--5 critical facts. Do not over-list.
  \item Keep each fact one short, declarative sentence.
  \item \textbf{STRICT NO-ANSWER-LEAK RULE}: Every listed fact must be DIRECTLY OBSERVABLE from the image (a label you can read, a line/region you can see, a count you can make). DO NOT include any of the following, no matter how true:
    \begin{itemize}[leftmargin=1.2em, labelsep=0.4em, itemsep=1pt, topsep=2pt, parsep=0pt]
      \item the ground-truth answer itself or any restatement,
      \item the result of a derivation (e.g., ``the maximum occurs at X'', ``the side equals Y'', ``$f(x)$ has Z roots''),
      \item properties that require solving the problem to know (e.g., ``triangle ABC is isosceles'' when isoscelesness is the conclusion, not a marked label).
    \end{itemize}
    If a candidate fact is something a solver PROVES rather than READS, drop it.
  \item Sanity check before emitting: for each fact, ask ``could a person who has NOT solved this problem look at the image and verify this fact in seconds?'' If no, the fact is leaking the solution; remove it.
\end{enumerate}

\smallskip
Return only one JSON object, no markdown, no extra text:

\begin{Verbatim}
{
  "gold_facts": [
    {"id": "F1", "fact": "...", "criticality": "critical", "fact_type": "..."},
    ...
  ]
}
\end{Verbatim}
\end{promptbox}

\paragraph{(P5) Per-fact PRD scoring.}
For each gold fact, we ask Gemini-3-Flash to independently decide whether it is (i) explicitly \emph{stated} in the response and (ii) \emph{instrumentally used} as a load-bearing premise in the derivation.

\begin{promptbox}
\footnotesize
You are an expert evaluator for multimodal mathematical reasoning. Your goal is to detect Perception-Reasoning Disconnect (PRD): cases where the model \emph{states} a correct visual fact but its \emph{derivation of the final answer} does not actually rely on that fact (drops it, substitutes a different value, or contradicts it).

\smallskip
\textbf{Question}\\
\texttt{\{PROBLEM\}}

\smallskip
\textbf{Ground-truth answer}\\
\texttt{\{GROUND_TRUTH\}}

\smallskip
\textbf{Gold visual facts} (anchor for evaluation; pre-extracted from image and solution)\\
\texttt{\{GOLD_FACTS_JSON\}}

\smallskip
\textbf{Model response} (verbatim, do NOT split or reformat)
\begin{Verbatim}
---
{MODEL_RESPONSE}
---
\end{Verbatim}

\smallskip
\textbf{Critical Distinction}.\\
Two questions you must keep STRICTLY SEPARATE:
\begin{enumerate}[leftmargin=2em, label=(\Alph*), labelsep=0.4em, itemsep=2pt, topsep=2pt, parsep=0pt]
  \item ``Did the model STATE this visual fact?'' \textbf{Step 1.} Narrow scope: only counts an explicit verbal mention or paraphrase of the fact in the response.
  \item ``Did the model's DERIVATION of the final answer ACTUALLY USE this fact as a load-bearing premise (i.e., the stated value/relation is plugged into a calculation, equation, or logical deduction that propagates to the answer)?'' \textbf{Step 2.} Strictly stronger than (A): a fact that appears only as decoration, in an unrelated step, or that is later bypassed by a different value DOES NOT count as load-bearing.
\end{enumerate}
A model can mention a fact and then ignore it, contradict it, or substitute a different value when computing.

\smallskip
\textbf{Calibration example} (study before answering).\\
Suppose Gold fact F1 = ``central angle AOB is labeled 140 in the diagram'', and the model writes: ``I observe that angle AOB is 140. To find angle ACB, I recall that the inscribed angle is typically half of 90, which is 45. So ACB = 45.''
Correct judgment for F1:
\begin{itemize}[leftmargin=1.2em, labelsep=0.4em, itemsep=1pt, topsep=2pt, parsep=0pt]
  \item \textbf{Step 1 (STATEMENT)}: \texttt{stated_in_response=true}, \texttt{stated_correctly=true}, \texttt{statement_quote="angle AOB is 140"}.
  \item \textbf{Step 2 (INSTRUMENTAL USE)}: \texttt{instrumental_in_derivation=false} (140 is mentioned, then ignored), \texttt{value_consistency="na"}, \texttt{contradicted_in_derivation=true} (``half of 90, which is 45'' implicitly uses a different angle than 140), \texttt{derivation_quote="half of 90, which is 45"}.
\end{itemize}
This is a textbook PRD: stated correctly but not load-bearing in the derivation.

\smallskip
\textbf{Evaluation Procedure}.

\textit{Step 1 -- STATEMENT (narrow scope: explicit mention only).} For EACH gold fact, output:
\begin{itemize}[leftmargin=1.2em, labelsep=0.4em, itemsep=1pt, topsep=2pt, parsep=0pt]
  \item \texttt{"id"}, \texttt{"stated_in_response"} (true\,|\,false),
  \item \texttt{"stated_correctly"} (true\,|\,false\,|\,\texttt{"na"}; \texttt{"na"} only if not stated),
  \item \texttt{"statement_quote"}: short verbatim quote, \texttt{""} if not stated.
\end{itemize}

\textit{Step 2 -- INSTRUMENTAL USE (load-bearing in derivation).} INDEPENDENT of Step 1; trace the derivation and decide whether the fact's value/relation actually carries through to the answer.
\begin{itemize}[leftmargin=1.2em, labelsep=0.4em, itemsep=1pt, topsep=2pt, parsep=0pt]
  \item \texttt{"id"}, \texttt{"instrumental_in_derivation"} (true\,|\,false): true iff the stated value/relation is plugged into a calculation, equation, or deductive step that leads to the final answer; false if it is decoration, never referenced again, or a different value is used instead.
  \item \texttt{"value_consistency"} (\texttt{"consistent"}\,|\,\texttt{"substituted"}\,|\,\texttt{"na"}): consistent if numeric/symbolic value used in derivation matches the stated one; substituted if a different value is used; na if non-numeric or \texttt{instrumental_in_derivation=false}.
  \item \texttt{"contradicted_in_derivation"} (true\,|\,false): true if any derivation step overrides, ignores-and-substitutes, or otherwise contradicts the stated fact.
  \item \texttt{"derivation_quote"}: short verbatim quote of the derivation step where the fact is used/substituted/contradicted, or \texttt{""}.
\end{itemize}

\textit{Step 3 -- Reasoning quality note (audit only; one sentence).} Briefly describe any logic or arithmetic errors. NOT used in P/U computation; recorded for analysis.

\textit{Step 4 -- Final answer.}
\begin{itemize}[leftmargin=1.2em, labelsep=0.4em, itemsep=1pt, topsep=2pt, parsep=0pt]
  \item \texttt{"predicted_answer"}: model's final answer as it appears, \texttt{""} if missing.
  \item \texttt{"answer_matches_ground_truth"}: true\,|\,false\,|\,\texttt{"unclear"} (semantic match allowed).
  \item \texttt{"answer_match_reason"}: one sentence.
\end{itemize}

\smallskip
\textbf{Output} (single JSON object, no markdown, no extra text):

\begin{Verbatim}
{
  "fact_statement": [
    {"id": "F1", "stated_in_response": true, "stated_correctly": true, "statement_quote": "..."}
  ],
  "fact_instrumental_use": [
    {"id": "F1", "instrumental_in_derivation": true, "value_consistency": "consistent", "contradicted_in_derivation": false, "derivation_quote": "..."}
  ],
  "reasoning_quality_note": "...",
  "predicted_answer": "...",
  "answer_matches_ground_truth": true,
  "answer_match_reason": "..."
}
\end{Verbatim}
\end{promptbox}

\end{document}